\documentclass[submission,copyright,creativecommons]{eptcs}
\providecommand{\event}{ICLP 2025} % Name of the event you are submitting to

\usepackage{iftex}
\usepackage{amsfonts}
\usepackage{listings}
\usepackage{numprint}
\usepackage[inline]{enumitem}
\usepackage[all]{xy}
\usepackage{xcolor}
\usepackage{times}
\usepackage{soul}
\usepackage{url}
\usepackage{hyperref}
\usepackage[utf8]{inputenc}
\usepackage[small]{caption}
\usepackage{graphicx}
\usepackage{amsmath}
\usepackage{amsthm}
\usepackage{algorithm}
\usepackage{algpseudocode}
\usepackage[switch]{lineno}
\usepackage{booktabs}
\usepackage{subcaption}
\usepackage{fixltx2e}
\MakeRobust{\Call}
\usepackage{forest}
\usetikzlibrary{angles}
\newtheorem{example}{Example}

\title{\CoolToolName: An Expert System for Danish Traffic Law}

\author{Luís Cruz-Filipe
\institute{IT University of Copenhagen\\ Copenhagen, Denmark}
\email{luic@itu.dk}
\and
Jonas Vistrup
\institute{University of Southern Denmark\\
Odense, Denmark}
\email{vistrup@imada.sdu.dk}
}

\theoremstyle{definition}

\lstdefinelanguage{law}{
    basicstyle=\small\ttfamily,
    keywordstyle=\upshape\bfseries,
    keepspaces=true,
    numbers=none,
    numbersep=8pt,
    breaklines=true,
    tabsize=2,
    keywords={},
    morecomment=[f][\itshape][0]{///},
}

\lstdefinelanguage{rules}{
    basicstyle=\small\ttfamily,
    keywordstyle=\upshape\bfseries,
    keepspaces=true,
    stepnumber=1,
    numbersep=8pt,
    breaklines=true,
    frame=lines,
    tabsize=2,
    backgroundcolor=\color{background},
    keywords={},
    escapeinside={\#}{)},
    literate={~} {$\sim$}{1},
    literate={\\S} {\S}{1},
    otherkeywords={<-,/\,\/},
    morecomment=[f][\itshape][0]{///},
}
\newcommand{\rules}[1]{\lstinline[language=rules]{#1}}

\definecolor{background}{HTML}{EEEEEE}
\definecolor{darkgreen}{rgb}{0,.5,0}
\definecolor{darkred}{rgb}{.8,0,0}

\newcommand{\CoolToolName}{\textsc{færdXel}}
\newcommand{\eoe}{\hspace*\fill$\triangleleft$}

\def\titlerunning{FærdXel}
\def\authorrunning{L. Cruz-Filipe \& J. Vistrup}
\begin{document}

\maketitle
\begin{abstract}
We present \CoolToolName, a tool for symbolic reasoning in the domain of Danish traffic law.
\CoolToolName{} combines techniques from logic programming with a novel interface that allows users to navigate through its reasoning process, thereby ensuring the system's explainability. Towards the goal of better understanding the value of \CoolToolName{}, two evaluations of the system have been performed:
(1)~An empirical evaluation showing that for a selection of court cases, the conclusions of \CoolToolName{} align with those of Danish judges.
(2)~A qualitative evaluation from legal experts indicating that this work has potential to become a foundation for real-world AI tools supporting professionals in the Danish legal sector.
\end{abstract}

\section{Introduction}
The release of ChatGPT in late 2022 sparked a global desire to integrate data-driven Artificial Intelligence (AI), based on sophisticated modern machine learning techniques, into as many domains as possible.
However, some domains seem resistant to integrating this type of AI into their workflow.
For some domains, machine learning is not a suitable approach, and for others, there is an established scepticism to the technology.

The domain of law exhibits both types of resistance.
First, in many legal systems, there is a mandate for an explanation of judgments~\cite{ml_law_req}, and it is unclear from precedent whether the explanation has to be the specific reason for the judgment or just a possible reason behind the judgment.
It can be argued that machine learning techniques, in the form of generative and deep neural networks, can only provide the latter.
Second, legal systems are traditionally based on discussion and argumentation, which clashes with AI systems that only provide definitive answers without justification.

One possible approach to overcome these issues is to revert to more traditional forms of AI, in the form of expert systems.
Expert systems provide symbolic reasoning using a knowledge base and a theoretically understood, sound inference algorithm.
They also can provide a trace of their inference process, allowing us to scrutinize the reasoning behind its answers and, in some sense, argue with the system.
However, the knowledge base must typically be tailored for the specific application in mind, so the system can reason within the domain of that application similar to how an expert might reason.
This makes building expert systems very time-consuming.
Another known limitation is that expert systems typically suffer from an interface problem, and using them often requires dedicated training -- limiting their widespread usage~\cite{no_new_expert_systems}.

\paragraph{This work.}
We present \CoolToolName,
an expert system designed to assist legal experts working with cases regarding possible violations of Danish traffic law.
Specifically, \CoolToolName{} looks for arguments supporting that a defendant has broken specific paragraphs of the Danish traffic law, both taking into account the law itself and similar cases that may be relevant.
The goal of \CoolToolName{} is not to make any decisions, but rather to provide meaningful and extensive input to those agents who will actually make the decisions -- in the same way that a legal aid would.
\CoolToolName{} is based on classic principles from logic programming.
Its knowledge base is manually written in an enriched version of Datalog, and it uses SLD-resolution as its inference system.
Every fact and rule in \CoolToolName's knowledge base comes with a schematic translation into natural language. \CoolToolName{} answers questions by creating SLD-refutations, which are used to build a tree-like structure of possible explanations for each answer.
\CoolToolName{} also provides an interface for navigating through these explanations in a user-friendly way using the provided translations.
Thus, users can examine \CoolToolName{}'s arguments and conclusions without understanding its internal language -- this knowledge is only required to enter new information to \CoolToolName, e.g., to edit facts about a specific case.

\CoolToolName{} was developed as a first step in an interdisciplinary collaboration to understand the potential usage of expert systems in the legal domain, at a time when data-driven AI (and especially generative AI) is opening up new possibilities.
Ultimately, our goal is to develop systems that can interact fully in natural language, and reason over a variety of domains.
We chose to focus initially on traffic law because it is a simple domain, with clear-cut rules, making the logic formulation technically simpler -- and allowing us to focus quicker on understanding how best to create and provide explanations.

\paragraph{Structure of the paper.}
Section~\ref{sec:related} contains the relevant related work for our development.
In Section~\ref{sec:back}, we describe the language and inference system of \CoolToolName, and how they differ from standard tools.
Section~\ref{sec:translate} describes in more detail the process of converting laws into logical rules.
The mechanism underlying the creation of explanations by \CoolToolName{} is described in Section~\ref{sec:explain}, and Section~\ref{sec:interface} focuses on the system's interface.
We have also undertaken both a small, qualitative evaluation of potential social value of \CoolToolName{} and an empirical evaluation of the correctness of its answer, both of which we describe in Section~\ref{sec:eval}.
Finally, Section~\ref{sec:concl} concludes and points out directions for future work.
The system implementation is available at \href{https://doi.org/10.5281/zenodo.15626892}{https://doi.org/10.5281/zenodo.15626892}.

\section{Related Work}\label{sec:related}
Research into expert systems and applications flourished in the 1980s including applications to the legal domain.
However, this direction of research slowed down in the 2000s, and since 2010, to the best of our knowledge, there have been few publications in the domain of expert systems for law~\cite{no_new_expert_systems}.
An interesting exception, from 2016, is an article on an expert system for Islam inheritance law~\cite{expert_law}.

Development of expert system applications for the legal domain go as far back to 1977's \textit{TAXMAN}~\cite{TAXMAN}, an expert system for answering questions about US taxation for corporate reorganization.
This work paved the way for several subsequent expert system, such as:
\textit{CHIRON}~\cite{CHIRON}, for planning actions that could decrease income tax according to US law;
\textit{Split up}~\cite{Split_up}, which dealt with Australian law on property division after divorce;  
and \textit{ASHSD-II}~\cite{ASHSD}, focusing on US law on property division after divorce.

All of these applications focus on very specific subdomains within already specific legal domains: 
In domain of US taxation, \textit{TAXMAN} on the subdomain of corporate reorganization and \textit{CHIRON} on the subdomain of income tax.
Reversely, both \textit{Split up} and \textit{ASHSD-II} on the subdomain of property division, but in different overarching domains, as \textit{Split up} deals with Australian divorce law and \textit{ASHSD-II} with US divorce law.
In contrast, \CoolToolName{} reasons on the whole of Danish traffic law, instead of being restricted to a subdomain like parking laws or driving laws.
The tool is also designed such that additional laws can be added to its knowledge.
 
Other applications have aimed at a broader approach by allowing for reasoning within arbitrary legal domains.
\textit{SHYSTER}~\cite{shyster} is an expert system for providing and arguing similarities and differences between cases.
It provided a framework for inputting knowledge about any legal case in any domain.
In contrast to our work, however, \textit{SHYSTER} did not contain knowledge about laws and was unable to reason about legal consequence.
This was rectified by combining \textit{SHYSTER} with a knowledge base.
The resulting system \textit{SHYSTER-MYCIN}~\cite{shyster_mycin} was thus able to reason using concrete laws, but again at the price of restricting its scope -- in this case, to the subdomain of copyright law within Australian property law.
In contrast, \CoolToolName{} is able to reason on the entirety of Danish Traffic law.

Research on expert system applications in other domains has also been lacking in the last decade due to the increasing success of machine learning based systems.
A notable exception is the medical diagnostic domain, where research into expert system applications has seen a low but steady output, demonstrate the utility of symbolic reasoning domains where the interpretability of results is critical, an important property within law.
Examples include systems for diagnosing chronic heart failure using logic programming~\cite{Chen16} and for diagnosing ischemic heart disease using the results of certain tests~\cite{medical1}.
More recently, an expert system provided non-experts with common diagnoses of their ankle problems~\cite{medical2}. The user would answer a series of questions and the system would either reach a conclusion or decide that a doctor should be consulted.
A questionnaire was also used to diagnose mental problems~\cite{medical3}.

Since the release of ChatGPT, there has also been a significant increase in AI based commercial tools to assist the daily work of legal experts~\cite{legalAIguide}.  
To the best of our knowledge, all recently released commercial AI legal tools are based on generative AI, primarily LLMs (Large Language Models)~\cite{harvey,ironclad,lexis,genie,cocounsel}. 
LLMs are able to generate arguments to a much wider degree than expert systems, since they can provide additional arguments based on the context and meaning of words~\cite{legalLLMinterpretation}.
However, these LLMs are trained on such large datasets that these sets likely contain undesired bias, which is transferred to the LLM during training~\cite{LLMgenderbias}. 
This bias can be introduced when the LLM is interpreting facts, deriving conclusions or creating justifications.

To remove undesired bias, the LLM will often go through rounds of post-training called alignment~\cite{alignmentguide,alignmentmethods}, e.g., reinforcement learning from human feedback~\cite{deepseek}, but it is currently unclear if these methods removes bias or only removes some expressions of bias \cite{aligmentfake}.
In fact, it could well be the case that tools for law based on LLMs could provide well-hidden biased answers to legal queries:
an LLM, which sequentially generates its output, could first generate a biased answers to a legal question and then generate a justification for said answer, selectively generating certain non-biased justifications, while leaving out the biased justifications for the answer, as the alignment has reduced the likelihood of the biased justification being presented. 
In that case, biased answers would be harder to detect after alignment, leading to undesired effects.
\CoolToolName{} avoids this issue by generating full explanations for answers before deducing answers and subsequently show them to the user.

\section{Background}\label{sec:back}
This section covers the necessary knowledge needed to understand the remainder of our work.
We discuss the logical language Datalog, which we use to represent the knowledge inherent in Danish traffic law, and briefly summarize the key concepts of SLD-resolution, the inference algorithm used by \CoolToolName{}.
Lastly, we provide a brief description of the Danish court system.

\subsection{Datalog}
\emph{Datalog} is the negation-free and function-free fragment of Prolog~\cite{datalog}.
The basic formulas of Datalog are called \emph{atoms}.
Each atom describes a relation between objects.
All atoms are on the form of $P(t_1,t_2,\ldots,t_n)$ where $P$ is a \emph{predicate} describing the type of relation and $t_1$ through $t_n$ are all \emph{terms} given as arguments for the predicate $P$.
A term can either be a \emph{constant} or a \emph{variable}.
A constant denotes a specific known object, and a variable denotes some unknown object.
A \emph{substitution} is a function mapping variables to terms, and the result of \emph{applying} a substitution to a formula is computed by simultaneously replacing each variable by its value according to the substitution.
The result of applying a substitution $\theta$ to a formula $\alpha$ is written~$\alpha\theta$.

A Datalog \emph{program} is a set of \emph{rules} (also called clauses).
Each rule is in the format $$\alpha \leftarrow \beta_1, \beta_2, \ldots, \beta_n$$
where $\alpha$ and $\beta_1$ through $\beta_n$ are atoms.
The semantic understanding of the rule is if $\beta_1$ through $\beta_n$ (the premises) holds then $\alpha$ (the conclusion) holds.
If $n=0$ then the rule simply states that $\alpha$ holds, and is called a \emph{fact}.

\subsection{SLD-resolution}
SLD-resolution is an algorithm to compute \emph{answers} to \emph{queries} over Datalog programs.
A query is a set of atoms, and an answer is a substitution of the variables in the query which makes the query a logical consequence of the program.

SLD-resolution uses \emph{SLD-derivations}, which work on \emph{goals}.
A goal is a formula of the form $\alpha_1,\ldots,\alpha_n$.
The basic ingredients for constructing SLD-derivations are \emph{unifiers} and \emph{resolution steps}.

A substitution $\theta$ \emph{unifies} two formulas $\alpha$ and $\beta$ if $\alpha\theta$ is the same as $\beta\theta$.
Finding whether such a substitution exists for given $\alpha$ and $\beta$ is known to be decidable, and there is a standard \emph{unification algorithm} that computes a specific unifier of any two formulas, if one exists.

Resolution chooses one atom $\alpha$ in a goal $G$ and a rule $r$ from the program such that $\alpha$ can be unified with the conclusion of $r$.
The result is a new goal $G'$, which is obtained by applying the unifier of $\alpha$ and the head of $r$ to both $G$ and $r$, and replacing $\alpha\theta$ in the result with the premise of $r\theta$.
(In particular, the selected atom is simply removed if $r$ is a fact.)
An SLD-derivation is a finite sequence of goals $G_0,\ldots,G_n$ where $G_0$ is the query and $G_{i+1}$ is obtained from $G_i$ by a resolution step.

An SLD-derivation also computes a substitution, which is obtained by composing the substitutions computed at each resolution step and restricting the result to the variables in the query.
If $G_n$ is a goal with no atoms, the SLD-derivation is called an \emph{SLD-refutation}.
Soundness of SLD-resolution guarantees that the substitution computed by an SLD-refutation starting with a goal $\leftarrow Q$ is an answer to the query $Q$, and completion of SLD-resolution guarantees that every answer to $Q$ can be obtained as a composition $\theta\sigma$, where $\theta$ is computed by an SLD-refutation with starting goal $\leftarrow Q$.

A more detailed discussion of SLD-resolution can be found in e.g.~\cite{Lloyd1984}.

\subsection{Specifics of the Danish court system}
In the Danish court system, the vast majority of cases is settled directly by sitting judges.
Defendants are only allowed to present their case in front of a jury in extreme criminal scenarios where the prosecution is demanding a prison sentence of more than 6 years.
These cases typically involve murder, rape, brutal robbery or extensive arson.

Since the judges are the arbitrators, court cases typically take on a procedural tone, and are usually predominantly technical.
Therefore, an expert system designed to be used in this setting should focus on the technical aspects of the law -- both what legislation says, and what can be inferred from similar cases.
This is also where many resources in the Danish legal system are used: both judges and lawyers are assisted by teams who gather information about relevant laws and cases, and help developing possible argumentations.

\section{Translation from Law to Rules}\label{sec:translate}
Reasoning systems are usually built from two components: a \emph{knowledge base}, containing information about the reasoning domain, and a \emph{reasoning engine} that can derive conclusions from the knowledge base.
In this section we focus on how \CoolToolName's knowledge base of 2579 rules was constructed.
Our goal was that this knowledge base should both have internal consistency, and be updatable in the future. Therefore we present the overarching structure of the translation from natural knowledge to \CoolToolName{}'s system knowledge.

Laws are typically formulated in the form of rules that must be upheld; however, legal argumentation is more often focused on whether someone has broken the law – this is also the task that \CoolToolName{} is expected to address.
To this end, we view the law as specifying pairs containing a situation and a prohibited action in that situation.
For each paragraph of the Danish traffic law, we formalize in Datalog both the situation the law is specifying, and the prohibited action.

Formalization can be done in a myriad of ways, so to ensure a consensus between rules in the knowledge base, some general principles have been developed and adhered to.
The main principle is one of `reification', which is the philosophical idea of describing abstract concepts as objects. 
In the context of \CoolToolName{} we first define objects, either physical or abstract, through single argument predicates, then use generic predicates named by function words (\rules{Has}, \rules{On}, \rules{With}) to describe the relation between the objects, and end by using specific predicates with few arguments (at most 4) to describe relations typically relevant only to the paragraph under consideration.

In addition, there is some information hidden in the context, and the applicable case law.
To ensure the inclusion of this information, we use a standard reference~\cite{faerdselMedKommentarer} containing an in-depth understanding of Danish traffic law and relevant case law as provided by a group of legal experts -- this is also an information source used by professionals.
Furthermore, to capture the knowledge of the law for events with causes and effects, the system must understand that something can be true at on time, yet false at another time.
Therefore all predicates $P(t_1,\ldots,t_n)$ have a time component $\tau$.
The system has two ways of expressing the time component of a predicates, either specifying the relation holds at time $\tau$ by adding it as another argument at the end, $P(t_1,\ldots,t_n,\tau)$, or by omitting it, $P(t_1,\ldots,t_n)$, to specify that the relation holds at all times.\footnote{Syntactically, this makes our language similar to Temporal Datalog~\cite{Motik}. However, we do not treat the temporal argument in any special way, so we can still remain in the realm of pure Datalog.}
These steps can be summarized, for each part of the law, as:
\begin{enumerate}
    \item[1)] Identify the situation and prohibited action using standard reference~\cite{faerdselMedKommentarer} and relevant case law.
    \item[2)] Formulate as much of the situation and action using previously defined predicates.
    \item[3)] Try to formulate remaining parts of the situation and action using function words between abstract or non-abstract objects.
    \item[4)] Formulate any remaining parts as specific predicates. 
\end{enumerate}

Often the law specifies similar situations where the same action is prohibited, or similar actions that are prohibited in the same situation.
Since each situation/action combo is translated into a rule, this leads to a combinatorial explosion where a single part of a paragraph in some cases generate over 50 rules that only slightly differ.
To make it easier to write such rules and minimize the risk of errors in translation, we include some syntactic sugar and allow premises of rules to include non-atomic formulas using disjunctions (\rules{\\\/}) and conjunctions (\rules{/\\}).
These formulas are then converted into Conjunctive Normal Form (CNF) using standard algorithms, yielding standard Datalog rules -- so that the knowledge base is still pure Datalog.

\begin{example}
\S4.1 of the Danish traffic law can be translated into the following English:
\begin{lstlisting}[language=law]
Road Users must follow the instructions given by traffic signs, 
  markings on road, traffic lights or a similar method. 
\end{lstlisting}
We use the steps for translating to derive a Datalog translation of \S4.1.
Step~1) Using the reference material~\cite{faerdselMedKommentarer}, the situation become a road user on a road with a traffic sign, marking, traffic light or similar, which are providing an instruction. The action is the road user not following the instruction.
Step~2) Let us assume \rules{BrokenLaw} and \rules{RoadUser} are previously defined predicates that can be reused.
Step~3) We can find the following objects: \rules{Instruction}, \rules{Road}, \rules{Sign}, \rules{Marking} and \rules{TrafficLight}. We can also identify the following function words: \rules{Has} and \rules{On}\mbox{}Step~4) As we are lacking a way to specify similarity and not following an instruction, the following specific predicates are defined:
\rules{SimilarTo} and \rules{NotFollowing}.
Using the language we have specified, $\S4.1$ can be translated into:
\begin{lstlisting}[language=rules]
BrokenLaw(P,\S4.1,T) <- RoadUser(P,T), Instruction(I), Road(R), 
  Sign(Z) \/ Marking(Z) \/ TrafficLight(Z) \/ SimilarTo(Z,a_sign)
  \/ SimilarTo(Z,road_marking) \/ SimilarTo(Z,traffic_light),
  Has(Z,I), On(P,R,T), On(Z,R), NotFollowing(P,I,T)
\end{lstlisting}
where \rules{P} is a person, \rules{I} an instruction, \rules{R} a road, \rules{Z} the object giving \rules{I} and \rules{T} is the time component.
\eoe
\end{example}

For presentation, it is essential that all predicates can be translated into natural language.
Therefore, \CoolToolName{}'s knowledge base also includes a set of definitions of natural-language variants for each predicate.
This is the only place where temporal arguments get special treatment, as they are handled uniformly for all predicates.

\begin{example}
The predicates used to implement \S4.1 are associated with the following natural language translations.
\begin{lstlisting}[language=rules,numbers=none]
RoadUser(P):P is a road user; Instruction(I):I is an instruction; 
Road(R):R is a road; Sign(S):S is a sign; Marking(M):M is a marking; 
TrafficLight(Z):Z is a trafficlight; On(A,B):A is on B; Has(A,B):A has B;
NotFollowing(P,I):P isn't following I; BrokenLaw(P,L):P has broken L;
\end{lstlisting}
Temporal arguments are omitted in these translations.
If a predicate includes a temporal argument \rules{T}, then \rules{at} \rules{T} is appended to its translation.
\eoe
\end{example}

\section{Inference System and Explainability}\label{sec:explain}
In this section, we discuss how the SLD-refutations created by the reasoning engine of \CoolToolName{} are compactly stored to provide explanations for the system's conclusions.

\begin{figure}
    \centering
    \includegraphics[width=\linewidth]{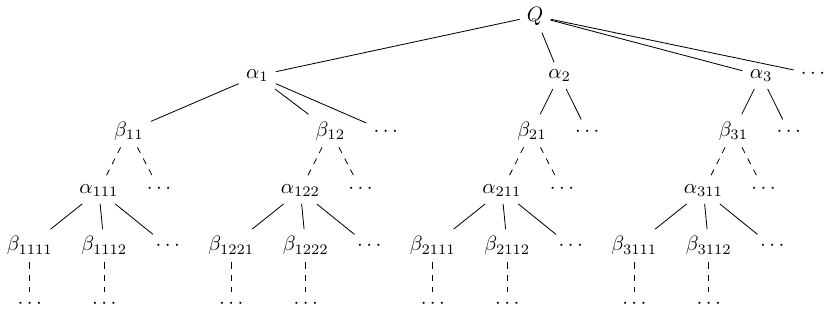}
    \caption{Explanation in \CoolToolName{}. $\alpha$-nodes are explanation-nodes and $\beta$-nodes are reasons-nodes.}
    \label{fig:explain_tree}
\end{figure}

Each explanation is stored in a tree format, as in Figure~\ref{fig:explain_tree}, where each node is an atom, and the root node is the original query.
All other nodes are separated into two types: explanations and reasons.
The type of nodes alternate between layers, i.e., the child of an explanation-node is a reason-node and vice versa. 
The children of the root node are explanation-nodes.

For each explanation-node $\alpha'$ with parent node $\beta'$, there exists a clause $\alpha \leftarrow \beta_1, \beta_2, \ldots, \beta_n$, such that $\beta'\theta = \alpha\theta = \alpha'$ for some substitution $\theta$, i.e., the parent node $\beta'$ unifies with the head of the clause $\alpha$ resulting in $\alpha'$. Its children are the reason-nodes $\beta_1\theta$, $\beta_2\theta$, \ldots, $\beta_n\theta$. 

An explanation-node should be interpreted as \emph{an} explanation for why the parent node's atom is valid and its children's atoms should be interpreted as the atoms that \emph{all} must be valid for an explanation to be valid  (this structure is a a variant of the classic AND/OR-tree). 

The tree is constructed using only valid SLD-refutations, which ensures that the tree is finite and that the atom of every node is valid \footnote{Validity, when the atom contains variables, requires that at least one assignment of the variables is valid}. 

\begin{figure}
    \centering
    \includegraphics[width=\linewidth]{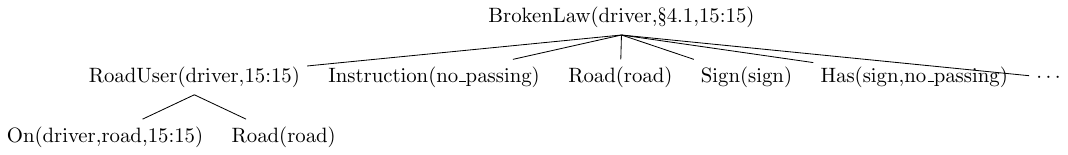}
    \includegraphics[width=\linewidth]{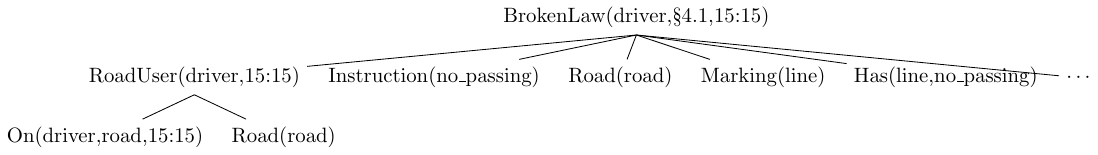}
    \includegraphics[width=\linewidth]{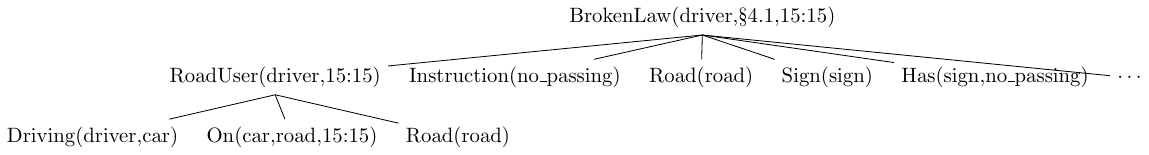}
    \includegraphics[width=\linewidth]{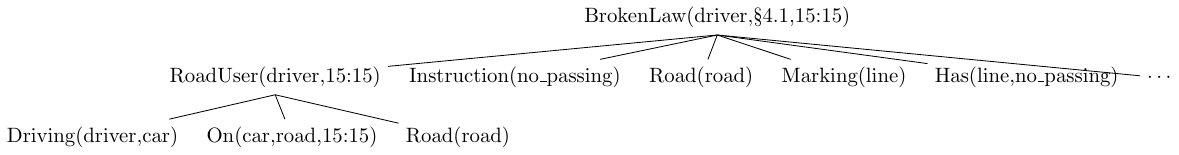}
    \caption{Four proof trees created by querying \rules{BrokenLaw(P,X,T)} on Example~\ref{ex:case}.}
    \label{fig:prooftree}
\end{figure}

\begin{figure}[b]
    \centering
    \includegraphics[width=\linewidth]{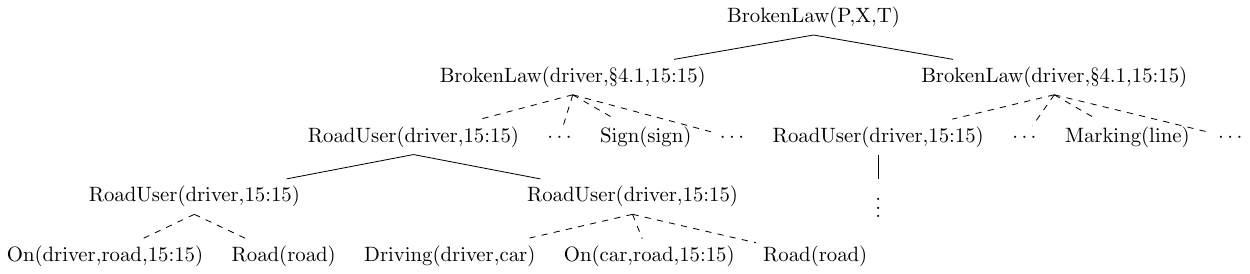}
    \caption{Explanation generated by \CoolToolName{} for query  \rules{BrokenLaw(P,X,T)} on Example~\ref{ex:case}.}
    \label{fig:explaintree}
\end{figure}

\begin{example}
To illustrate the process we now introduce a case that uses the paragraphs of the law encoded in the previous section.
This case is also used for showcasing the interface in Section~\ref{sec:interface}.

The case regards a person overtaking a patrol officer on a road where passing is not allowed~\cite{domsag}.
The overtaking occurs at 15:15, and there are both a sign and road markings stating that passing is not allowed.
For privacy reasons, certain details of the case are anonymized, so the person is referred to as the \rules{driver}, their car as \rules{car}, and the road as \rules{road}.

The facts of the cases can be encoded as:
\begin{lstlisting}[language=rules]
Driving(driver,car), Road(road), Sign(sign), On(sign,road), 
Marking(line), Instruction(no_passing), Has(sign,no_passing), 
Has(line,no_passing), On(line,road), On(car,road,15:15), 
On(driver,road,15:15), NotFollowing(driver,no_passing,15:15)
\end{lstlisting}

In addition to the encoding of $\S4.1$ from the previous section, assume our knowledge base also includes two rules derived from $\S2.25$ of the Danish traffic law, which define the predicate \rules{RoadUser}:
\begin{lstlisting}[language=rules]
RoadUser(P,T) <- On(P,R,T), Road(R)
RoadUser(P,T) <- Driving(P,C), On(C,R,T), Road(R)
\end{lstlisting}
To find out if any laws have been broken, we send the query \rules{BrokenLaw(P,X,T)} to \CoolToolName{}, which first generates all possible SLD-refutations using SLD-resolution.
In this case, four refutation are created together with their proof tree (Figure~\ref{fig:prooftree}).
These proof trees are then combined to create an explanation (Figure~\ref{fig:explaintree}).
\label{ex:case}
\end{example}


\section{Interface}\label{sec:interface}

We now showcase the interface of \CoolToolName{} from the perspective of a user, using the case introduced in the previous section (Section~\ref{sec:explain}).
Our focus is on displaying explanations in an understandable way and allowing users to navigate them. Currently our prototype has only a minimalistic interface providing these capabilities.

We do not include actual screenshots, but instead redrawings of the interface. This is done both to exacerbate details and because the actual interface is in Danish.
For illustration, we use the case presented in the previous section.

\begin{figure}[t]
\centering
\begin{subfigure}{0.48\textwidth}
    \includegraphics[width=\textwidth]{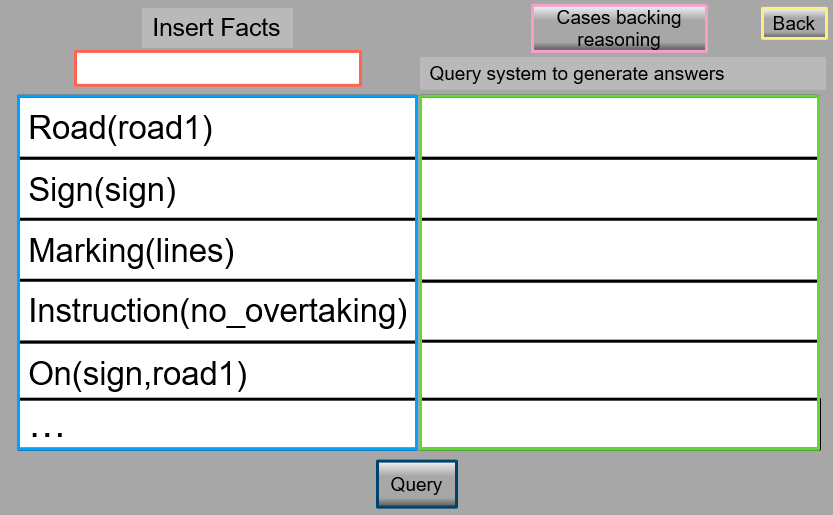}
    \caption{Interface after case-facts have been entered.}
    \label{fig:zero}
\end{subfigure}
\hfill
\begin{subfigure}{0.48\textwidth}
    \includegraphics[width=\textwidth]{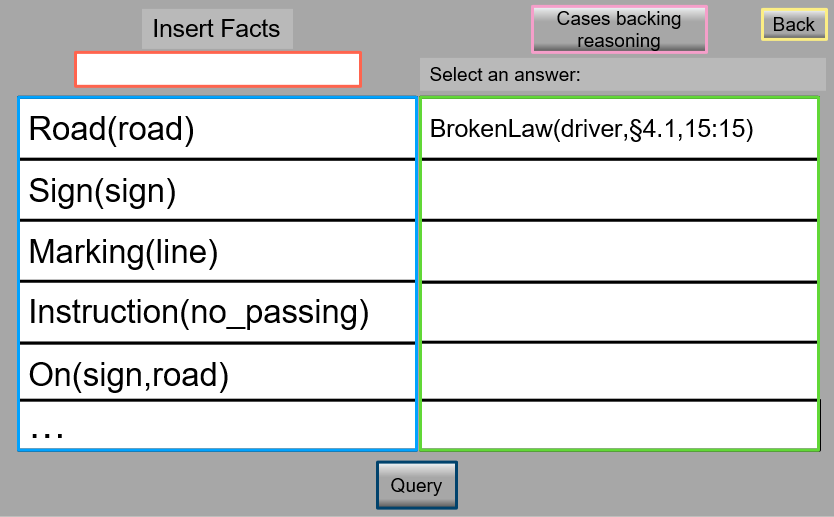}
    \caption{Interface after the system has been queried.}
    \label{fig:first}
\end{subfigure}
\hfill
\begin{subfigure}{0.48\textwidth}
    \includegraphics[width=\textwidth]{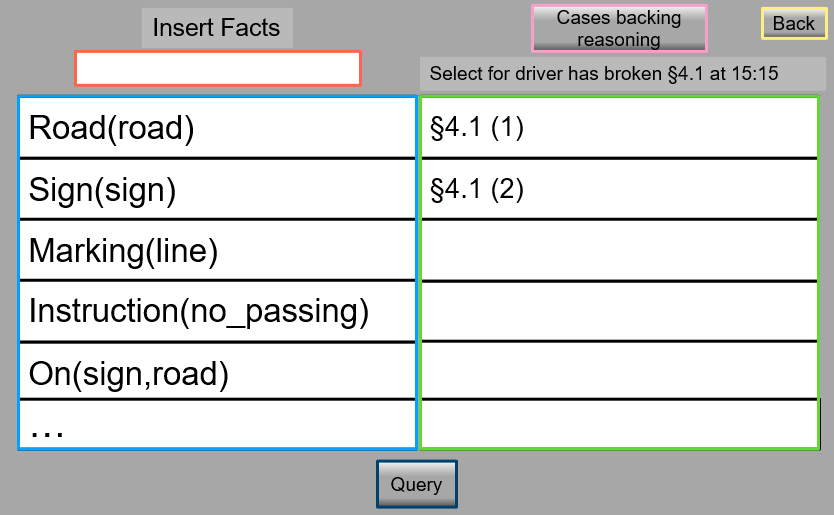}
    \caption{Interface after an answer has been selected.}
    \label{fig:second}
\end{subfigure}
\hfill
\begin{subfigure}{0.48\textwidth}
    \includegraphics[width=\textwidth]{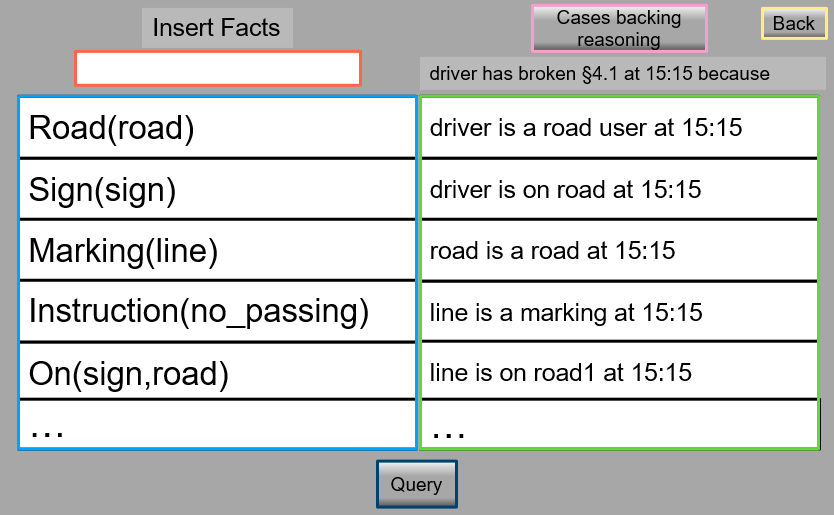}
    \caption{Interface after an explanation has been selected.}
    \label{fig:third}
\end{subfigure}
\caption{Illustrations of the interface. The three horizontal dots at the bottom of a display box informs the user that additional items can be found by scrolling.}
\label{fig:figures}
\end{figure}

Figure~\ref{fig:zero} shows the interface after all facts about the case have been entered and Figure~\ref{fig:first} after the reasoning algorithm has been run. 
The left side of the interface displays the facts entered for the case, and the right side shows answers and explanations to queries.
All facts are entered using the red box under ``Insert Facts'', and must be written in Datalog using the predefined predicates. Entered facts are shown in the blue box on the left side, and can be removed again by clicking them.
The reasoning algorithm runs the query \rules{BrokenLaw(P,X,T)} on the entered facts when the dark blue ``Query'' button is pressed, after which the answers that have been found are shown in the green box on the right side.

Clicking on an answer selects it, and replaces the right side of the window with a list of immediate explanations (the child nodes of the root in the tree generated by the query, Figure~\ref{fig:second}).
At this point the yellow ``Back'' button in the top right corner becomes active, allowing the user to reverse the last selection performed in the green box.

Figure~\ref{fig:third} shows the selected immediate explanation for the previously selected answer.
The green box now displays the premises of the rule used to derive the answer.
Pressing the ``Cases backing reasoning'' button shows references to the paragraphs (or cases, or both) from which the rule is derived.
Further clicking on the immediate explanations shown will in turn exhibit their own immediate explanations (child nodes on the tree).
Eventually, every branch of the tree will end with some leaf nodes, which are exactly the facts from the case (or from the knowledge base).

As shown, the interface allows for good traversability and understandability of explanations.

\section{Evaluation and Target Users}\label{sec:eval}

Two aspects of \CoolToolName{} have been evaluated: the correctness of its conclusions and the value it might add to the Danish legal system. 
To justify the choice of these evaluations, we first describe the intended use of the tool.

\CoolToolName{} is intended to be used on Danish traffic law cases, by having a user enter facts about the case, allowing the tool to providing sound arguments for and against law-breaches.
Its purpose is to assist legal professionals, and provide argumentations in situations that relies on law and legal praxis that has been thoroughly decided. 
\CoolToolName{} is not intended to be a judge of facts or of judgements -- this task is performed by the court system.

Our first evaluation intends to ascertain whether \CoolToolName{} is correct, in the sense that its conclusions are valid given the facts entered as input.
Our second evaluation looks into the more empirical assessment of whether \CoolToolName{} indeed adds value to professionals working within the Danish legal system, and identifies concrete areas of applicability.

\paragraph{Evaluation of correctness.}
The correctness of \CoolToolName{}'s conclusions was tested by applying the system to on-going court cases, and analysing the differences between the breaches of law deduced by \CoolToolName{} and those reached by the court.
The system was applied by having an expert user entering the facts of the case in the Datalog format.
All facts were entered during the trial, and before the defendant was given a verdict.
These findings are summarized in Table~\ref{tab:eval1}.

\begin{table}[t]
\centering
\begin{tabular}{p{0.75cm}p{5cm}p{5cm}}
         \toprule
         Case & \CoolToolName{}'s  & Court's \\
         Nr.     & Conclusion         & Conclusion \\
         \midrule
         1 & § 3 &§ 3 \\
         2 & \phantom{}{\color{darkred}None} &§ 82\\
         3 & \phantom{}{\color{darkgreen}§ 3}, § 15 & § 15\\
         4 & § 3, § 15 & § 3, § 15\\
         5 & § 42, {\color{darkred}§ 44} & § 42\\
         6 & § 3, § 15, § 54&§ 3, § 15, § 54\\
         7 & § 42, § 53 &§ 42, § 53\\
         8 & \phantom{}{\color{darkred}§ 56} &§ 117\\
         9 &§ 3, § 15, § 54&§ 3, § 15, § 54\\
         10 & § 3, § 9, {\color{darkgreen}§ 15}, § 53&§ 3, § 9, § 53\\
         \bottomrule
\end{tabular}

\caption{Conclusions on what laws were breached according to \CoolToolName{} and the courts respectively for 10 different traffic law trial.
    Red color indicates incorrect conclusion, and green color indicates correct conclusion but not what the defendant was on trial for.}
\label{tab:eval1}
\end{table}

The system was tested on ten court cases.
For five of the cases (1, 4, 6, 7 and 9) there were no difference between the conclusions of \CoolToolName{} and the conclusions of the courts. 

For two of the cases (3, 10), \CoolToolName{} concluded an additional breach of law beyond those reached by the courts.
In both cases, the defendant admitted to additional facts during trial that led to those conclusions, but the Danish courts are only allowed to judge on breaches brought forth prior to trial, and those additional breaches were not among them.
In a real-life setting, these additional facts would not be introduced into \CoolToolName{}, and these additional breaches would not be derived.

In case 5, which was about speeding on the highway, \CoolToolName{} incorrectly deduced that the defendant broke \S 44, concerning which vehicles are allowed on the highway, because the expert user forgot to enter that the vehicle the defendant drove was a car.
After adding this fact, the system again derived the correct answer.

A similar situation occurs with case 8, about a driver with a revoked license.
The system incorrectly deduced a breach of \S 56 (driving before acquiring a license) instead of \S 117 (driving after getting your license suspended).
This was due to the expert user incorrectly inputting that the defendant did not have their license instead of having their license suspended, and fixing the input again led to the correct conclusion.

The remaining case (2) deals with a chair falling from a trailer, and the defendant was found to be innocent by \CoolToolName{}.
Due to a mistake, an information hidden in case law was not included in the translation of the law to rules.
Therefore the system incorrectly assumed that for \S 82, concerning unsecured goods, a person needed to be in possible danger, when actually it only needs to be likely for a person to be where the goods fell.
This is the only case where \CoolToolName{}'s knowledge was incorrect, but in a similar way to humans, the knowledge of the system can be updated, ensuring it will not make the same mistake again.
After updating the rule, \CoolToolName{} was able to derive the right conclusion.

Since \CoolToolName{} provides explanations for all conclusions, it is easy to refute false positives, such as in case 5 and partly in case 8.
However, since the system can not explain why it did not deduce a specific conclusion, it is harder to debunk false negatives.
Specifically, detecting the error in case~2 requires knowing the relevant legislation.

In the context of using \CoolToolName{} as a tool in the courts, the fact that false positives are easier to detect than false negatives is aligned with the principle in Western court systems that it is preferable to let a guilty man go free than convict an innocent one (presumption of innocence).

\paragraph{Evaluation of social value.}
To evaluate the social value of \CoolToolName{}, the system has been presented to a group of Danish legal experts, with the goal of getting feedback on the system and its potential uses.
Initially these experts were sceptical, but after understanding the mechanisms of the system better, they showed excitement for the project.
They believe that \CoolToolName{} and similar tools can provide great help in a number of areas.
They also identified bottlenecks for usability, pointing out that only an expert can input facts in Datalog -- an instance of a well-known general issue with expert systems~\cite{GM18}.
They also identified four possible groups of end-users who would benefit from \CoolToolName{}: Traffic police, prosecutors, citizen and judges.

\emph{Traffic police officers} conduct multiple traffic stops a day, each of which has to be supported by a valid legal reason. These officers have a quick-reference guide to assist them in conducting valid stops, however as the law is quite complex, the guide is a simplification of the actual legal current praxis.
This creates false positives that need to be solved in court, adding to the workload of the legal system.
Replacing this quick-reference guide by a tool such as \CoolToolName{} would likely help reduce the number of false positives, due to \CoolToolName's better understanding of Danish traffic law.
However, this application requires an improvement to the current interface, as any viable system would need to be as fast to use as the existing guide.

\emph{Prosecutors} examine hundreds of potential cases per month and need to make quick decisions about which ones are solid enough to merit the effort of prosecution.
They typically rely on assistants (very often law students) to suggest different possible arguments, whose strength they then evaluate. 
One known problem is that these assistants occasionally make hard-to-detect typos that effect the validity of the case.
Using \CoolToolName{} as an additional assistant could heavily reduce the workload of prosecutors, while also reducing the amount of minor errors.

\emph{Common citizens} involved in a legal issue often have to analyse the cost-benefits of hiring a (likely expensive) lawyer for a specific case.
This results citizens often representing themselves in smaller cases. 
To mitigate the disadvantages of the citizen not having a lawyer, the law requires that both the judge and prosecutor fairly defends the interests of the citizen, adding an additional work load to both judges and prosecutors. 
The legal system as a whole could benefit immensely from an online system that would help citizens understand why they are being accused of breaking the law and whether it is meaningful to fight the accusation in court.

Finally, \emph{judges} need to make rulings on cases and decide on sentencing, with limited time to do so.
They also rely on assistants who look for similar cases and present arguments for heavier or lighter sentencing.
In this domain, fairness is the dominating criterion meaning that \CoolToolName{} would unlikely be useful as a replacement of assistant -- but it could still very useful as an additional source of thorough arguments.

The end-user group of judges was further explored by allowing a judge to interact with the system before and after a trial, and getting their feedback on needed improvement. The judge heavily emphasised that reasoning about punishment, would be the most useful aspect for a judge.

All findings on use-cases are summarized in Table~\ref{tab:eval2}.

\begin{table}[t]
\centering
    \begin{tabular}{p{4cm}p{5cm}p{5cm}}
    \toprule
       \emph{End-User}  & \emph{Benefits} & \emph{Requirements of \CoolToolName{}} \\
    \midrule
       Traffic \mbox{Police}    & Identifying law-breaking drivers & Fast and mostly automatic\\ \midrule
       Prosecutors              & Providing legal arguments  & Fast to use\\ \midrule
       Citizens                 & Answering legal questions & Usable by non-expert\\ \midrule
       Judges                   & Providing counter-arguments & Reason about punishment\\
    \bottomrule
    \end{tabular}
    \caption{Summarized feedback from legal experts about possible end-user for \CoolToolName\, the way they could benefit, and the improvement needed for that type of user.}
    \label{tab:eval2}
\end{table}

\section{Conclusions and Future Work}\label{sec:concl}
The entire Danish traffic law has been manually encoded into \CoolToolName, allowing it to answer queries regarding legal cases using SLD-resolution and provide explanations behind its answers. The implementation is available at \href{https://doi.org/10.5281/zenodo.15626892}{https://doi.org/10.5281/zenodo.15626892}.
Our empirical evaluation highlights the potential that can arise out of the collaboration between \CoolToolName{} and a human user, with our panel of  legal expert identifying multiple suitable application domains.
We plan to explore these applications and whether we can improve the quality of the explanations provided by \CoolToolName{}.

As a next step, we want to extend \CoolToolName's reasoning capabilities to the domain of sentencing.
Danish traffic law states a range of punishment for a given crime and defines aggravating and mitigating circumstances, but it is up to the judge to decide what the exact fine or prison sentence should be.
We plan to explore how to apply of Fuzzy Logic to this task, by viewing `aggravating' and `mitigating' as Fuzzy terms in the domain of possible punishments.
A similar approach, from which we draw inspiration, has been applied to Polish punishment law~\cite{fuzzy}.

Lastly, to improve the usability of \CoolToolName, we are exploring how to use large language models in a reliable way to convert written natural language about a case into the Datalog format that the tool requires as input.
We believe that this could be the way to develop a user-friendly interface that could make \CoolToolName{} easily accessible to non-experts.
The major challenge is to ensure that the translation is correct, requiring the user to confirm that the model's ``understanding'' of the input is the right one.

% \newpage
\bibliographystyle{eptcs}
\bibliography{bibliography}

\begin{thebibliography}{10}
\providecommand{\bibitemdeclare}[2]{}
\providecommand{\surnamestart}{}
\providecommand{\surnameend}{}
\providecommand{\urlprefix}{Available at }
\providecommand{\url}[1]{\texttt{#1}}
\providecommand{\href}[2]{\texttt{#2}}
\providecommand{\urlalt}[2]{\href{#1}{#2}}
\providecommand{\doi}[1]{doi:\urlalt{https://doi.org/#1}{#1}}
\providecommand{\eprint}[1]{arXiv:\urlalt{https://arxiv.org/abs/#1}{#1}}
\providecommand{\bibinfo}[2]{#2}

\bibitemdeclare{article}{expert_law}
\bibitem{expert_law}
\bibinfo{author}{Alaa~N. \surnamestart Akkila\surnameend} \& \bibinfo{author}{Samy S.~Abu \surnamestart Naser\surnameend} (\bibinfo{year}{2016}): \emph{\bibinfo{title}{Proposed Expert System for Calculating Inheritance in Islam}}.
\newblock {\slshape \bibinfo{journal}{World Wide Journal of Multidisciplinary Research and Development}} \bibinfo{volume}{2}(\bibinfo{number}{9}), pp. \bibinfo{pages}{38--48}.

\bibitemdeclare{article}{ml_law_req}
\bibitem{ml_law_req}
\bibinfo{author}{Adrien \surnamestart Bibal\surnameend}, \bibinfo{author}{Michael \surnamestart Lognoul\surnameend}, \bibinfo{author}{Alexandre \surnamestart de~Streel\surnameend} \& \bibinfo{author}{Benoît \surnamestart Frénay\surnameend} (\bibinfo{year}{2020}): \emph{\bibinfo{title}{Legal requirements on explainability in machine learning}}.
\newblock {\slshape \bibinfo{journal}{Artifcial Intelligence and Law}} \bibinfo{volume}{29}, pp. \bibinfo{pages}{149--169}, \doi{10.1007/s10506-020-09270-4}.

\bibitemdeclare{inproceedings}{Chen16}
\bibitem{Chen16}
\bibinfo{author}{Zhuo \surnamestart Chen\surnameend} (\bibinfo{year}{2016}): \emph{\bibinfo{title}{Automating Disease Management Using Answer Set Programming}}.
\newblock In \bibinfo{editor}{Manuel \surnamestart Carro\surnameend}, \bibinfo{editor}{Andy \surnamestart King\surnameend}, \bibinfo{editor}{Neda \surnamestart Saeedloei\surnameend} \& \bibinfo{editor}{Marina~De \surnamestart Vos\surnameend}, editors: {\slshape \bibinfo{booktitle}{Technical communications of ICLP}}, {\slshape \bibinfo{series}{OASIcs}}~\bibinfo{volume}{52}, \bibinfo{publisher}{Schloss Dagstuhl - Leibniz-Zentrum f{\"{u}}r Informatik}, pp. \bibinfo{pages}{22:1--22:10}, \doi{10.4230/OASICS.ICLP.2016.22}.

\bibitemdeclare{article}{legalLLMinterpretation}
\bibitem{legalLLMinterpretation}
\bibinfo{author}{Andrew \surnamestart Coan\surnameend} \& \bibinfo{author}{Harry \surnamestart Surden\surnameend} (\bibinfo{year}{2025}): \emph{\bibinfo{title}{Artificial Intelligence and Constitutional Interpretation}}.
\newblock {\slshape \bibinfo{journal}{University of Colorado Law Review}} \bibinfo{volume}{413}, \doi{10.2139/ssrn.5018779}.

\bibitemdeclare{misc}{harvey}
\bibitem{harvey}
\bibinfo{author}{Counsel~AI \surnamestart Corporation\surnameend} (\bibinfo{year}{2025}): \emph{\bibinfo{title}{Harvey AI}}.
\newblock \urlprefix\url{https://www.harvey.ai/}.

\bibitemdeclare{misc}{domsag}
\bibitem{domsag}
\bibinfo{author}{Danmarks \surnamestart Domstole\surnameend} (\bibinfo{year}{2022}): \emph{\bibinfo{title}{Domsdatabasen}}.
\newblock \bibinfo{howpublished}{\url{https://domsdatabasen.dk/\#sag/2526/3008}}.
\newblock \bibinfo{note}{[Accessed 23-09-2024]}.

\bibitemdeclare{article}{medical2}
\bibitem{medical2}
\bibinfo{author}{Basel~Y. \surnamestart Elhabil\surnameend} \& \bibinfo{author}{Samy~S. \surnamestart Abu{-}Naser\surnameend} (\bibinfo{year}{2021}): \emph{\bibinfo{title}{An Expert System for Ankle Problems}}.
\newblock {\slshape \bibinfo{journal}{International Journal of Engineering and Information Systems (IJEAIS)}} \bibinfo{volume}{5}(\bibinfo{number}{4}), pp. \bibinfo{pages}{57--66}.

\bibitemdeclare{article}{no_new_expert_systems}
\bibitem{no_new_expert_systems}
\bibinfo{author}{Mohammad \surnamestart Farajollahi\surnameend} \& \bibinfo{author}{Vahid \surnamestart Baradaran\surnameend} (\bibinfo{year}{2024}): \emph{\bibinfo{title}{Expert System Application in law: A review of research and applications}}.
\newblock {\slshape \bibinfo{journal}{International Journal of Nonlinear Analysis and Applications}} \bibinfo{volume}{15}(\bibinfo{number}{8}), pp. \bibinfo{pages}{107--114}, \doi{10.22075/ijnaa.2023.31260.4596}.

\bibitemdeclare{article}{datalog}
\bibitem{datalog}
\bibinfo{author}{Herve \surnamestart Gallaire\surnameend}, \bibinfo{author}{Jack \surnamestart Minker\surnameend} \& \bibinfo{author}{Jean-Marie \surnamestart Nicolas\surnameend} (\bibinfo{year}{1984}): \emph{\bibinfo{title}{Logic and Databases: A Deductive Approach}}.
\newblock {\slshape \bibinfo{journal}{ACM Comput. Surv.}}, p. \bibinfo{pages}{153–185}, \doi{10.1145/356924.356929}.

\bibitemdeclare{article}{aligmentfake}
\bibitem{aligmentfake}
\bibinfo{author}{Ryan \surnamestart Greenblatt\surnameend}, \bibinfo{author}{Carson \surnamestart Denison\surnameend}, \bibinfo{author}{Benjamin \surnamestart Wright\surnameend}, \bibinfo{author}{Fabien \surnamestart Roger\surnameend}, \bibinfo{author}{Monte \surnamestart MacDiarmid\surnameend}, \bibinfo{author}{Samuel \surnamestart Marks\surnameend}, \bibinfo{author}{Johannes \surnamestart Treutlein\surnameend}, \bibinfo{author}{Tim \surnamestart Belonax\surnameend}, \bibinfo{author}{Jack \surnamestart Chen\surnameend}, \bibinfo{author}{David \surnamestart Duvenaud\surnameend}, \bibinfo{author}{Akbir \surnamestart Khan\surnameend}, \bibinfo{author}{Julian \surnamestart Michael\surnameend}, \bibinfo{author}{S{\"{o}}ren \surnamestart Mindermann\surnameend}, \bibinfo{author}{Ethan \surnamestart Perez\surnameend}, \bibinfo{author}{Linda \surnamestart Petrini\surnameend}, \bibinfo{author}{Jonathan \surnamestart Uesato\surnameend}, \bibinfo{author}{Jared \surnamestart Kaplan\surnameend}, \bibinfo{author}{Buck
  \surnamestart Shlegeris\surnameend}, \bibinfo{author}{Samuel~R. \surnamestart Bowman\surnameend} \& \bibinfo{author}{Evan \surnamestart Hubinger\surnameend} (\bibinfo{year}{2024}): \emph{\bibinfo{title}{Alignment faking in large language models}}.
\newblock {\slshape \bibinfo{journal}{CoRR}}, \doi{10.48550/arXiv.2412.14093}.

\bibitemdeclare{article}{GM18}
\bibitem{GM18}
\bibinfo{author}{Graham \surnamestart Greenleaf\surnameend}, \bibinfo{author}{Andrew \surnamestart Mowbray\surnameend} \& \bibinfo{author}{Philip \surnamestart Chung\surnameend} (\bibinfo{year}{2018}): \emph{\bibinfo{title}{Building sustainable free legal advisory systems: Experiences from the history of {AI} {\&} law}}.
\newblock {\slshape \bibinfo{journal}{Comput.\ Law Secur.\ Rev.}} \bibinfo{volume}{34}(\bibinfo{number}{2}), pp. \bibinfo{pages}{314--326}, \doi{10.1016/J.CLSR.2018.02.007}.

\bibitemdeclare{article}{medical1}
\bibitem{medical1}
\bibinfo{author}{Mohammad~Shahadat \surnamestart Hossain\surnameend}, \bibinfo{author}{Mohammad~A. \surnamestart Haque\surnameend}, \bibinfo{author}{Rashed \surnamestart Mustafa\surnameend}, \bibinfo{author}{Razuan \surnamestart Karim\surnameend}, \bibinfo{author}{Hirak~Chandra \surnamestart Dey\surnameend} \& \bibinfo{author}{Md. \surnamestart Yusuf\surnameend} (\bibinfo{year}{2016}): \emph{\bibinfo{title}{An Expert System to Assist the Diagnosis of Ischemic Heart Disease}}.
\newblock {\slshape \bibinfo{journal}{International Journal of Integrated Care}} \bibinfo{volume}{16}, p.~\bibinfo{pages}{31}, \doi{10.5334/ijic.2974}.

\bibitemdeclare{misc}{ironclad}
\bibitem{ironclad}
\bibinfo{author}{Inc \surnamestart Ironclad\surnameend} (\bibinfo{year}{2025}): \emph{\bibinfo{title}{Ironclad}}.
\newblock \urlprefix\url{https://ironcladapp.com/product/ironclad-for-legal/}.

\bibitemdeclare{article}{medical3}
\bibitem{medical3}
\bibinfo{author}{Somay \surnamestart Jain\surnameend}, \bibinfo{author}{Mukul \surnamestart Aggarwal\surnameend}, \bibinfo{author}{Yash \surnamestart Singhal\surnameend} \& \bibinfo{author}{Apri~Dwi \surnamestart Lestari\surnameend} (\bibinfo{year}{2023}): \emph{\bibinfo{title}{An expert system on diagnosis of mental diseases}}.
\newblock {\slshape \bibinfo{journal}{Journal of Soft Computing Exploration}} \bibinfo{volume}{4}, pp. \bibinfo{pages}{53--58}, \doi{10.52465/joscex.v4i1.100}.

\bibitemdeclare{misc}{lexis}
\bibitem{lexis}
\bibinfo{author}{\surnamestart LexisNexis\surnameend} (\bibinfo{year}{2025}): \emph{\bibinfo{title}{Lexis+AI}}.
\newblock \urlprefix\url{https://www.lexisnexis.com/en-us/products/lexis-plus-ai.page}.

\bibitemdeclare{article}{alignmentguide}
\bibitem{alignmentguide}
\bibinfo{author}{Yang \surnamestart Liu\surnameend}, \bibinfo{author}{Yuanshun \surnamestart Yao\surnameend}, \bibinfo{author}{Jean-Francois \surnamestart Ton\surnameend}, \bibinfo{author}{Xiaoying \surnamestart Zhang\surnameend}, \bibinfo{author}{Ruocheng \surnamestart Guo\surnameend}, \bibinfo{author}{Hao \surnamestart Cheng\surnameend}, \bibinfo{author}{Yegor \surnamestart Klochkov\surnameend}, \bibinfo{author}{Muhammad~Faaiz \surnamestart Taufiq\surnameend} \& \bibinfo{author}{Hang \surnamestart Li\surnameend} (\bibinfo{year}{2024}): \emph{\bibinfo{title}{Trustworthy LLMs: a Survey and Guideline for Evaluating Large Language Models' Alignment}}.
\newblock {\slshape \bibinfo{journal}{arXiv preprint}}, \doi{10.48550/arXiv.2308.05374}.

\bibitemdeclare{book}{Lloyd1984}
\bibitem{Lloyd1984}
\bibinfo{author}{John~W. \surnamestart Lloyd\surnameend} (\bibinfo{year}{1984}): \emph{\bibinfo{title}{Foundations of Logic Programming}}.
\newblock \bibinfo{publisher}{Springer}.

\bibitemdeclare{inproceedings}{fuzzy}
\bibitem{fuzzy}
\bibinfo{author}{Michal \surnamestart Lower\surnameend} \& \bibinfo{author}{Monika \surnamestart Lower\surnameend} (\bibinfo{year}{2019}): \emph{\bibinfo{title}{Fuzzy Inference Model for Punishing a Perpetrator in a Judicial Process}}.
\newblock In \bibinfo{editor}{Ngoc~Thanh \surnamestart Nguyen\surnameend}, \bibinfo{editor}{Richard \surnamestart Chbeir\surnameend}, \bibinfo{editor}{Ernesto \surnamestart Exposito\surnameend}, \bibinfo{editor}{Philippe \surnamestart Aniorté\surnameend} \& \bibinfo{editor}{Bogdan \surnamestart Trawiński\surnameend}, editors: {\slshape \bibinfo{booktitle}{Procs.\ {ICCCI19}}}, \bibinfo{publisher}{Springer}, pp. \bibinfo{pages}{463--473}, \doi{10.1007/978-3-030-28377-3\_38}.

\bibitemdeclare{misc}{genie}
\bibitem{genie}
\bibinfo{author}{Genie~AI \surnamestart Ltd.\surnameend} (\bibinfo{year}{2025}): \emph{\bibinfo{title}{Genie AI}}.
\newblock \urlprefix\url{https://www.genieai.co/en-us}.

\bibitemdeclare{article}{TAXMAN}
\bibitem{TAXMAN}
\bibinfo{author}{L.~Thorne \surnamestart McCarty\surnameend} (\bibinfo{year}{1977}): \emph{\bibinfo{title}{Reflections on TAXMAN: An Experiment in Artificial Intelligence and Legal Reasoning}}.
\newblock {\slshape \bibinfo{journal}{Harvard Law Review}} \bibinfo{volume}{90}, pp. \bibinfo{pages}{837--893}.

\bibitemdeclare{inproceedings}{shyster_mycin}
\bibitem{shyster_mycin}
\bibinfo{author}{Thomas~A. \surnamestart O'Callaghan\surnameend}, \bibinfo{author}{James \surnamestart Popple\surnameend} \& \bibinfo{author}{Eric \surnamestart McCreath\surnameend} (\bibinfo{year}{2003}): \emph{\bibinfo{title}{SHYSTER-MYCIN: a hybrid legal expert system}}.
\newblock In: {\slshape \bibinfo{booktitle}{Procs.\ ICAIL '03}}, \bibinfo{publisher}{Association for Computing Machinery}, p. \bibinfo{pages}{103–104}, \doi{10.1145/1047788.1047814}.

\bibitemdeclare{article}{ASHSD}
\bibitem{ASHSD}
\bibinfo{author}{Kamalendu \surnamestart Pal\surnameend} \& \bibinfo{author}{John~A. \surnamestart Campbell\surnameend} (\bibinfo{year}{1997}): \emph{\bibinfo{title}{An application of rule-based and case-based reasoning within a single legal knowledge-based system}}.
\newblock {\slshape \bibinfo{journal}{SIGMIS Database}} \bibinfo{volume}{28}(\bibinfo{number}{4}), p. \bibinfo{pages}{48–63}, \doi{10.1145/277339.277344}.

\bibitemdeclare{book}{shyster}
\bibitem{shyster}
\bibinfo{author}{James \surnamestart Popple\surnameend} (\bibinfo{year}{1993}): \emph{\bibinfo{title}{SHYSTER: A Pragmatic Legal Expert System: PhD thesis}}.
\newblock \bibinfo{publisher}{Australian National Univ.}

\bibitemdeclare{misc}{cocounsel}
\bibitem{cocounsel}
\bibinfo{author}{Thomson \surnamestart Reuters\surnameend} (\bibinfo{year}{2025}): \emph{\bibinfo{title}{CoCounsel}}.
\newblock \urlprefix\url{https://www.thomsonreuters.com/en}.

\bibitemdeclare{inproceedings}{Motik}
\bibitem{Motik}
\bibinfo{author}{Alessandro \surnamestart Ronca\surnameend}, \bibinfo{author}{Mark \surnamestart Kaminski\surnameend}, \bibinfo{author}{Bernardo~Cuenca \surnamestart Grau\surnameend}, \bibinfo{author}{Boris \surnamestart Motik\surnameend} \& \bibinfo{author}{Ian \surnamestart Horrocks\surnameend} (\bibinfo{year}{2018}): \emph{\bibinfo{title}{Stream Reasoning in {Temporal Datalog}}}.
\newblock In \bibinfo{editor}{Sheila~A. \surnamestart McIlraith\surnameend} \& \bibinfo{editor}{Kilian~Q. \surnamestart Weinberger\surnameend}, editors: {\slshape \bibinfo{booktitle}{Procs.\ AAAI}}, \bibinfo{publisher}{{AAAI} Press}, pp. \bibinfo{pages}{1941--1948}, \doi{10.1609/aaai.v32i1.11537}.

\bibitemdeclare{inproceedings}{CHIRON}
\bibitem{CHIRON}
\bibinfo{author}{Kathryn~E. \surnamestart Sanders\surnameend} (\bibinfo{year}{1991}): \emph{\bibinfo{title}{Representing and reasoning about open-textured predicates}}.
\newblock In: {\slshape \bibinfo{booktitle}{Procs.\ ICAIL}}, \bibinfo{publisher}{Association for Computing Machinery}, p. \bibinfo{pages}{137–144}, \doi{10.1145/112646.112663}.

\bibitemdeclare{article}{legalAIguide}
\bibitem{legalAIguide}
\bibinfo{author}{Daniel \surnamestart Schwarcz\surnameend} \& \bibinfo{author}{Jonathan~H. \surnamestart Choi\surnameend} (\bibinfo{year}{2023}): \emph{\bibinfo{title}{AI Tools for Lawyers: A Practical Guide}}.
\newblock {\slshape \bibinfo{journal}{Minnesota Law Review Headnotes}} \bibinfo{volume}{1}, \doi{10.2139/ssrn.4404017}.

\bibitemdeclare{book}{faerdselMedKommentarer}
\bibitem{faerdselMedKommentarer}
\bibinfo{author}{Henrik \surnamestart Waaben\surnameend} \& \bibinfo{author}{Kirsten~Søndergaard \surnamestart Munck\surnameend} (\bibinfo{year}{2023}): \emph{\bibinfo{title}{Færdselsloven med kommentarer}}.
\newblock \bibinfo{publisher}{Djøf}.

\bibitemdeclare{inproceedings}{LLMgenderbias}
\bibitem{LLMgenderbias}
\bibinfo{author}{Yixin \surnamestart Wan\surnameend}, \bibinfo{author}{George \surnamestart Pu\surnameend}, \bibinfo{author}{Jiao \surnamestart Sun\surnameend}, \bibinfo{author}{Aparna \surnamestart Garimella\surnameend}, \bibinfo{author}{Kai-Wei \surnamestart Chang\surnameend} \& \bibinfo{author}{Nanyun \surnamestart Peng\surnameend} (\bibinfo{year}{2023}): \emph{\bibinfo{title}{``Kelly is a Warm Person, Joseph is a Role Model'': Gender Biases in {LLM}-Generated Reference Letters}}.
\newblock In \bibinfo{editor}{Houda \surnamestart Bouamor\surnameend}, \bibinfo{editor}{Juan \surnamestart Pino\surnameend} \& \bibinfo{editor}{Kalika \surnamestart Bali\surnameend}, editors: {\slshape \bibinfo{booktitle}{Procs.\ EMNLP}}, \bibinfo{publisher}{Association for Computational Linguistics}, pp. \bibinfo{pages}{3730--3748}, \doi{10.18653/v1/2023.findings-emnlp.243}.

\bibitemdeclare{article}{deepseek}
\bibitem{deepseek}
\bibinfo{author}{Chengen \surnamestart Wang\surnameend} \& \bibinfo{author}{Murat \surnamestart Kantarcioglu\surnameend} (\bibinfo{year}{2025}): \emph{\bibinfo{title}{A Review of DeepSeek Models' Key Innovative Techniques}}.
\newblock {\slshape \bibinfo{journal}{arXiv preprint}}, \doi{10.48550/arXiv.2503.11486}.

\bibitemdeclare{article}{alignmentmethods}
\bibitem{alignmentmethods}
\bibinfo{author}{Yufei \surnamestart Wang\surnameend}, \bibinfo{author}{Wanjun \surnamestart Zhong\surnameend}, \bibinfo{author}{Liangyou \surnamestart Li\surnameend}, \bibinfo{author}{Fei \surnamestart Mi\surnameend}, \bibinfo{author}{Xingshan \surnamestart Zeng\surnameend}, \bibinfo{author}{Wenyong \surnamestart Huang\surnameend}, \bibinfo{author}{Lifeng \surnamestart Shang\surnameend}, \bibinfo{author}{Xin \surnamestart Jiang\surnameend} \& \bibinfo{author}{Qun \surnamestart Liu\surnameend} (\bibinfo{year}{2023}): \emph{\bibinfo{title}{Aligning Large Language Models with Human: A Survey}}.
\newblock {\slshape \bibinfo{journal}{arXiv preprint}}, \doi{10.48550/arXiv.2307.12966}.

\bibitemdeclare{article}{Split_up}
\bibitem{Split_up}
\bibinfo{author}{J~\surnamestart Zeleznikow\surnameend} \& \bibinfo{author}{A~\surnamestart Stranieri\surnameend} (\bibinfo{year}{1998}): \emph{\bibinfo{title}{Split up: an intelligent decision support system which provides advice upon property division following divorce}}.
\newblock {\slshape \bibinfo{journal}{International Journal of Law and Information Technology}} \bibinfo{volume}{6}(\bibinfo{number}{2}), pp. \bibinfo{pages}{190--213}, \doi{10.1093/ijlit/6.2.190}.

\end{thebibliography}

\end{document}